# *Integrating wind variability to modelling wind-ramp events using a non-binary ramp function and deep learning models*

Russell Sharp[1, 2], Hisham Ihshaish[2], J. Ignacio Deza[2]

[1] *SP Energy Networks, St Vincent Crescent, Glasgow, G3 8LT, UK.*

[2] *Computer Science Research Centre (CSRC), Faculty of Environment and Technology, University of the West of England, Bristol, BS16 1QY, UK.*

## Abstract

*The forecasting of large ramps in wind power output known as ramp events is crucial for the incorporation of large volumes of wind energy into national electricity grids. Large variations in wind power supply must be compensated by ancillary energy sources which can include the use of fossil fuels. Improved prediction of wind power will help to reduce dependency on supplemental energy sources along with their associated costs and emissions. In this paper, we discuss limitations of current predictive practices and explore the use of Machine Learning methods to enhance wind ramp event classification and prediction. We additionally outline a design for a novel approach to wind ramp prediction, in which high-resolution wind fields are incorporated to the modelling of wind power.*

## 1. INTRODUCTION

In recent decades, renewable energy sources have received significant attention due to global concerns over climate change and carbon emissions. As part of their commitments to the 2016 UN Paris Agreement, industrialised countries of the EU have committed to supplying certain proportions of their energy demand using renewable sources by the year 2030. The focus of this project is on wind energy data from two specific members of the EU: France and Spain. In order to meet its climate change objectives, the French government has pledged to increase its installed wind energy capacity from 16 GW in 2021 to a minimum of 38.4 GW in 2028 (Abassi et al., 2016). Meanwhile, the Spanish government has submitted plans to increase its capacity from 28 GW in 2020 to 50.3 GW in 2030 (European Commission, 2019). The dataset from La Haute Borne wind farm, north-eastern France, will be reported for initial analysis.

Despite the benefits of wind power, its potential as an energy source is hindered by its inherent intermittency, owing to its direct correlation to the natural variability of the wind. Generation is further beset by sudden large ramps in wind power output known as ramp events. Generally speaking, ramp events represent large and fast variations in power output from a wind farm or portfolio of wind farms. They are driven by naturally occurring, sudden large increases (positive ramp) or decreases (negative ramp) in wind strength referred to as wind ramps. Ramp events are manifest as local events in a wind power time series, generally over a short period of up to a few hours (Cutler et al., 2007; Gallego-Castillo et al., 2015) (Fig. 1).

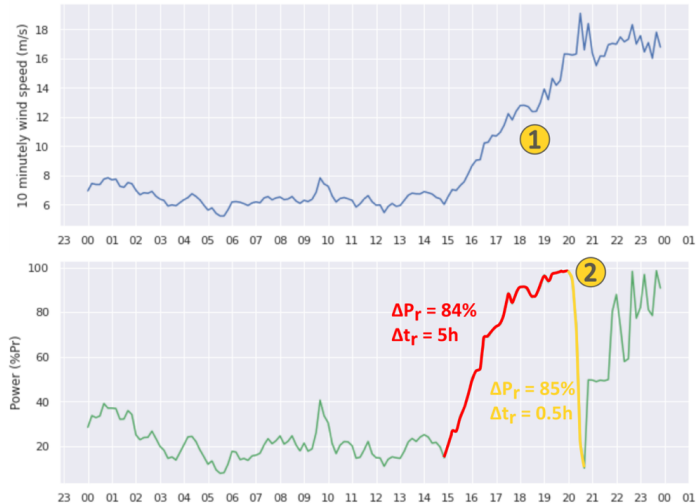

**Figure 1.** Example of two ramp events experienced on 13 January 2013 at the La Haute Borne wind farm, north-eastern France. Hour is plotted along the x-axis. Parameters employed to characterise ramps are $P_r$ = Rated Power and $t_r$ = time period over which the ramp takes place. Ramp direction is given by colour: red = ramp-up (positive), yellow = ramp-down (negative). The ramp-up event is driven by increasing wind speed (1). The ramp-down event in this instance is triggered as the farm's rated power is exceeded (2) and the turbines are shut-in to prevent damage. A detailed description of the dataset is presented in section 3.1.

The variability of wind power is a prominent obstacle for electricity Transmission System Operators (TSOs) towards incorporating commercial volumes into national power grids. TSOs must continually manage their power networks so that supply meets demand (Cutler et al., 2007). This is commonly done through the scheduling of ancillary reserves which consist of supplementary power sources that are flexible enough to adapt to variations in load and supply. This, however, carries additional costs and emissions.

Most electricity markets operate around a set of short-term procedures known as Day-Ahead (DA) operations that enable them to prepare for Real-Time (RT) energy dispatch operations. Part of the DA operations is the allocation of ancillary reserves: by a given deadline, market participants submit bids of the volumes of reserves that they can provide for the following day to the TSO. As RT operations approach, the TSO will perform a re-commitment procedure to account for any forced outages and the forecasted load on the grid (Monteiro et al., 2009; Martínez-Arellano, 2015).

Wind power forecasts then, are essential for the incorporation of wind energy into power networks over the DA timeframe. This is especially true during ramp events. From a TSO point of view, a large negative ramp in wind power could require a fast response from ancillary sources to maintain the supply-demand balance, whilst a large positive ramp might require that power flow be kept below planned maxima in constrained parts of the network (Cutler et al., 2007). As the amount of wind energy in a country's energy mix increases, so too does the importance of accurate wind power predictions. These forecasts are mainly characterised by time horizon, that is, the future time period spanned by the prediction. Time horizons are generally separated into three categories, depending on the market (Table 1).

Table 1. Categories of wind power forecast time horizon (after Martinez-Arellano, 2015).

| Category | Horizon |
| --- | --- |
| *Very short-term* | 2 – 4/9 hours |
| *Short-term* | 4/9 – 48/72 hours |
| *Medium-term* | 72 hours - 7 days |

Due to the increasing volume of wind power that is set to be incorporated into electricity markets by TSOs, and due to the DA operational timescales outlined above, short-term forecasts of wind power are currently an active area of research and form the horizon of interest in this research.

The aim of this study is to investigate potential ways to improve the characterisation and prediction of ramp events using Machine Learning methods. In order to achieve this, three main objectives are identified here and outlined further in the following sections: 1) Investigate wavelet signal transformation for the ramp event characterisation. It is anticipated that by using the WT, a ramp function is obtained which provides a continuous index related to ramp intensity at each time step of a given wind power time series. This in turn will enable the identification of ramp events often not captured by current binary ramp detection tools; 2) Study the feasibility of using Machine Learning models to learn the dynamics of the ramp function from wind power time-series; 3) Explore the use of prevailing wind direction to better account for misplacement errors in Numerical Weather Prediction (NWP) model outputs. Increasing the relative importance of upwind windspeed predictions is expected to reduce misplacement errors.

## 2. RELATED WORK

Machine Learning and time series forecasting techniques have been applied to the problem of wind power ramp event forecasting with some promising results (Ahmadi and Khashei, 2021; Gallego-Castillo et al., 2015; Sim and Yung, 2020; Wang et al., 2019; Yang et al., 2021 and references therein) but the field of study is still in its infancy.

There are two types of method for wind power forecasting: those based on time series analysis alone, and those based on a combination of time series analysis and Numerical Weather Prediction (NWP) model outputs. NWP models resolve a set of physical equations in order to estimate the dynamics of the atmosphere and output forecasted values of target variables to a 3D grid, but they do so at significant computational cost. A high-resolution NWP run of 0.5 km for example may take 48 hours to complete with limited resources, effectively rendering it useless for DA operations. Whilst TS+NWP prediction models typically out-perform TS approaches after a 3 to 6-hour time horizon (Giebel et al., 2011; Martinez-Arellano, 2015) the computational demands of NWPs exert challenging constraints on their deployment in any domain-specific application. Downscaling is a procedure that reduces computational cost by transforming NWP outputs from the low-resolution grids of NWP models to higher resolutions at specific physical locations of interest. It is commonly performed by statistical analysis of historic data to establish systematic relationships between NWP forecasts and measured observations (Cutler et al., 2007). Downscaling though, still depends on the interrogation and use of NWP datasets and carries with it associated computational and user expertise demands.

Most utilities require a short-term forecast of wind power, but one that is generated at relatively low computational cost and with relatively low user expertise requirements. Motivated by this, this study explores the possibility of a simplified wind power forecasting process based on Machine Learning methodologies.

### 2.1. Ramp event definition

Broadly, a ramp event is a large and rapid variation in wind power output (Cutler et al., 2007; Gallego-Castillo et al., 2015). However, the relative interpretation of 'large' and 'rapid' will differ according to the following factors:

a) The end use of any ramp forecasting model. For example, a wind farm operator interested in projected market penalties will be interested in different time scales as compared to an energy trader interested in instantaneous market demand and spot prices (Cutler et al., 2007; Martínez-Arellano, 2015).
b) The size of the wind farm/ portfolio. As an example, when defining ramps using an absolute power amplitude threshold value, a higher frequency of ramp events is likely to be observed as the installed capacity increases (Gallego-Castillo et al., 2015).

c) The cost function considered. For instance, costs of ancillary reserves and electricity market penalties (Gallego et al., 2013).

Ramp events can generally be identified and characterised considering the following features (Table 2):

**Table 2.** Ramp characterisation parameters used in the literature (after Gallego-Castillo et al., 2015).

| Term | Parameter | Description |
| --- | --- | --- |
| Magnitude | $\Delta P$ | Variation in power observed. |
| Duration | $\Delta t$ | Time period over which a variation takes place. |
| Ramp rate | $\Delta P / \Delta t$ | Variation divided by duration. Indicative of ramp intensity. |
| Timing | $t_0$ | Time instance of ramp event. Can be start or central time. |
| Direction | +/- | Increase/ ramp-up (+) or decrease/ ramp-down (-) in power output. |

Extracting ramp events from wind power time series, the parameters of table 2 can easily be analysed (e.g., Fig. 1). However, ramp forecasting usually entails the reverse; given certain characteristics or criteria, a forecaster must identify ramp events in order to determine underlying causes and create accurate predictive models. This establishes the need to set such criteria, or in other words: a ramp definition. Ramp event forecasting is a relatively immature research field. In the absence of a standard formal ramp event definition, the literature reports different characterisations depending on wind farm size and quantity (if aggregated), or the characteristics of the hosting power grid (Gallego-Castillo et al., 2015; Martínez-Arellano, 2015; Yang et al., 2021).

Most previous work has classified ramp events using a binary definition (Table 3). Binary definitions determine whether a ramp exists or not based on defined threshold values of magnitude and duration ($\Delta P_r$ and $\Delta t_r$ respectively, Fig. 1). This approach, however, has two major disadvantages. The first is that classification can become highly sensitive to the (often arbitrary) threshold values used. For instance, with $\Delta P_r$ set to 50%, an impactful change in power output of 49% may not be detected. The second is that a binary approach characterises all ramps as similar to one another, despite the fact that ramps with different characteristics are often observed. Ultimately, a binary definition restricts the forecaster from exploiting potential relationships between different ramp levels and continuous explanatory variables such as NWP outputs, SCADA data or meteorological tower measurements (Gallego et al., 2013). Table 4 summarises some previously used binary ramp event definitions and their limitations. It is worth noting that, despite the clear drawbacks of binary ramp event definitions, many recent works continue to use and refer to them (Table 3).

In order to overcome the drawbacks of the binary definition approach, Gallego et al. (2013) introduced the idea of using wavelet transform (WT) to characterise ramp events. The method requires the manual manipulation of only one input parameter (related to maximum ramp duration) which leaves model tuning in the hands of the end user. Using the WT methodology, a ramp function is obtained which provides a continuous index related to ramp intensity at each time step of a given wind power time series. Wavelet transform is adopted in this study (reported later in Section 3.2) to identify ramp events, thus adding the desired outcomes or 'labels' to the dataset that is introduced to the ML algorithms. A comprehensive review of wind power ramp forecasting was undertaken by Gallego-Castillo et al. in 2015. A summary of the relevant work in the field that has been carried out since its publication is provided in Table 3.

## 3. RESEARCH DESIGN AND METHODS

There are three main steps for short-term wind power prediction using NWP models: 1) downscaling, 2) conversion to power and 3) upscaling. Several downscaling approaches have been explored in published literature with varying degrees of success (Cutler et al, 2009; Gallegos-Castillo 2015; Martínez-Arellano, 2015). This study explores a novel downscaling approach to meteorological fields, and Machine Learning methods to provide feasible and comparably accurate wind ramp event prediction. The technical design for this proposal is shown in figure 2, and discussed throughout the remainder of this section.

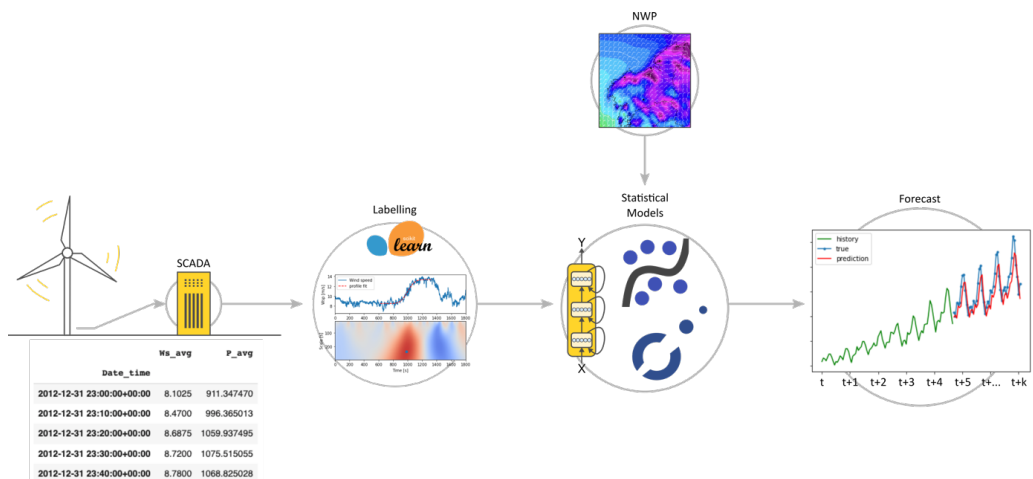

**Figure 2.** Flow chart of the methodology employed in this research. Ramps are identified ('labelled') in the raw data using *wavelet transform*. The family of ML algorithms is trained on the raw data, then used to forecast power output and associated ramp events. Using this methodology, NWP outputs may be used as an alternative data source (see text for explanation). Ramp function plot from Hannesdóttir and Kelly (2019a). NWP output from Xunta de Galicia (2021).

The learning methods on wind power time-series, as a univariate system, will include the Autoregressive Moving Average (ARMA) and its integrated and seasonal variants (ARIMA and SARIMA) to solve for seasonality and non-stationarity of the system. This is besides Recurrent Neural Networks, namely the LSTMs as well as Prophet time-series model of Facebook. The performance of the above models will be examined with respect to the accuracy of their ramp event identification and prediction. Additionally, a multivariate analysis of the system will include NWP high-resolution outputs so that wind power system can be analysed and subsequently predicted through wind field observations/assimilated values or predictions, instead of wind power historic data.

### 3.1. Dataset

The dataset used in this project is a Supervisory Control and Data Acquisition (SCADA) dataset collected from the La Haute Borne wind farm between December 2012 and January 2018. La Haute Borne is an 8,200-kW capacity onshore wind farm located in north-eastern France, centred at 48.45°N, 5.59°W. The farm is owned by the French energy and utilities company, Engie, and comprises four Senvion MM82 wind turbines. SCADA data generated from an array of sensors within each turbine is recorded in two large csv files (1.105 GB in total).

The files contain readings of each turbine's components such as rotor speed and gearbox bearing temperatures, grid information such as frequency and voltage, and, of most relevance to this study, physical information including wind speed, wind direction and active power. Table 5 provides a description of the variables used in this study.

### 3.2. Ramp characterisation

The ramp characterisation method used in this paper focusses on the DA TSO end-use case introduced in Section 1. During a ramp-down event a TSO must compensate for the loss of generation through the scheduling of ancillary re serves. During a ramp-up event the TSO may need to limit excessive wind power input to ensure grid safety (Martínez-Arellano, 2015).

**Table 3.** Main ramp forecasting literature reviewed[1].

| Author | Year | Model type | Definition | Algorithms/ methodology | Evaluation metrics | Horizons | Outputs | Location |
|---|---|---|---|---|---|---|---|---|
| Fernández et al. | 2013 | TS | Binary | Local Mahalanobis K-NN, Anisotropic Diffusion | CM | 3 h | Ramp occ. | ESP |
| Sevilan and Rajagopal | 2013 | TS | Binary | Dynamic Programming | CM | Varied | Power output | USA |
| Martínez-Arellano et al. | 2014 | NWP + TS | Non-binary | GP | CM, FIS score | 100 h | Power output | ESP |
| Martínez-Arellano | 2015 | NWP + TS | Non-binary | GP, Spatial fields | FIS score | 12-36 h | Power output | ESP |
| Ji et al. | 2015 | TS | Non-binary | WT, neuro-fuzzy network | n/a | n/a | Power output | n/a |
| Heckenbergerova et al. | 2016 | TS | Binary | PCA | Box plots | 4 h | Ramp occ. (prob) | USA |
| Gallego-Castillo et al. | 2016 | NWP + TS | Non-binary | Reproducing Kernel Hilbert Space | CRPS | 12 h | Power output (prob) | USA |
| Zha et al. | 2016 | TS | Binary | SVR, Swinging door algorithm | MAE, MAPE, RMSE, CC | 3.5-5.5 h | Power output | USA |
| Li et al | 2017 | NWP + TS | Binary | Gabor Filtering, BPNN | FA, RC, RMSE | 24 h | Power output | CAN |
| Dorado-Moreno et al. | 2017 | NWP + TS | Binary | Reservoir computing, Echo state networks | CM, Sensitivity, Specificity, GMS | 6-24 h | Ramp occ. | ESP |
| Taylor | 2017 | TS | Binary | AR logit models | Brier score | 1-2 h | Power output (prob) | GRC |
| Zhang et al. | 2017 | NWP + TS | Binary | Swinging Door Algorithm | POD, CSI, FBIAS | 1-6 h | Power output (prob) | USA |
| Liu et al. | 2017 | NWP + TS | Binary | Orthogonal Test, SVM | RC, FA, CSI | 0.5 to 24 h | Ramp occ. | CHN |
| Dorado-Moreno et al. | 2018 | NWP + TS | Binary | RNN, Echo State, Delay line reservoir | Geometric mean, CCR, AMAE | 6 h | Power output | ESP |
| Dhiman et al. | 2019 | TS | Binary | SVR, WT | AE, RMSE, MAE, CPU time | 0.5-3 h | Power output | ESP, USA, AUS, IND |
| Zhang et al. | 2019 | TS | Binary | NN, ARMA, WSR, PSO, KDE | MAE, MAPE, MSE, RMSE | 16 h | Wind speed | CHN, ESP |
| Liu et al. | 2019 | TS | Binary | KDE, Back Propagation Neural Network (BPNN) | RMSE | 4 h | Power output | CHN |
| Ouyang et al. | 2019 | NWP + TS | Binary | MLP | BIAS, MAE, RMSE, CM | 72 h | Power output | CHN |
| Fujimoto et al. | 2019 | NWP + TS | Binary | RF, RUS, ROS, SMOTE, Laplacian kernel function | RMSE, Precision, Recall, CSI | 0.5-48 h | Ramp occ., power output | JPN |
| Zhao et al. | 2020 | TS | Non-binary | Bayesian Network | CM | 30 min | Ramp occ. (prob) | CHN |
| Lyners et al. | 2020 | TS | Binary | Multi-parameter segmentation algorithm | PDFs | 25 h | Ramp occ. | ZAF |
| Ye et al. | 2020 | NWP + TS | Non-binary | Wave Division, Grey Wolf Optimizer, Fuzzy C-means clustering, LSTM | NMAE, NRMSE, ACR | 24-72 h | Power output | CHN |
| Cornejo-Bueno et al. | 2020 | NWP + TS | Binary | Extreme Learning Machine, SVR, SMOTE | ROC, CM | 6 h | Ramp occ. | ESP |
| Dorado-Moreno et al. | 2020 | NWP + TS | Binary | Multi-task Learning, Deep Neural Networks | Accuracy, Sensitivity, GMS | 6 h | Ramp occ. | ESP |
| Lyners et al. | 2021 | TS | Binary | Multi-parameter segmentation algorithm | POD, CSI, FBIAS, SR | n/a | Power output | ZAF |
| Hirata et al. | 2021 | TS | Non-binary | K-NN | MAE, Ignorance score | 1-12 h | Power output | Japan |
| Pichault et al. | 2021 | TS | Non-binary | Haar WT, Ramp Function | Statistical analysis | 1 h | Ramp characterisation | AUS |
| Couto et al. | 2021 | NWP | Binary | ANN | Statistical analysis | 24 h | Power output | PRT |
| Dhiman and Deb | 2021 | TS | Non-binary | TSVR, RFR, CNN, WT | RMSE | 10 min | Power output | UK, NLD, AUS |
| Zhou et al. | 2021 | TS | Binary | Generative Adversarial Network | MAE, MAPE, RMSE, FA, RC | 42 h | Power output | BEL, CHN |

**Table 4.** Binary ramp definitions used in the literature.
* Ramp-up is signified by $P_t < (Pt + \Delta t)$, ramp-down by $P_t > (Pt + \Delta t)$.
** Wind power can exhibit high variability over timescales shorter than typical ramp lengths. Therefore, ramp characterisation may become sensitive to noise. To overcome this issue, Bossavy et al. (2010) introduced the idea of a filtered signal.

| Reference | Definition | Description | Limitations |
|---|---|---|---|
| Kamath (2010)* | $P_t + \Delta t - P_t > P_{val}$ | A ramp event occurs if the magnitude of the change in the power signal between two time series observations exceeds a pre-set threshold. | Based on start and end values of $\Delta t$ so doesn't account for ramps that may occur during the interval. |
| Kamath (2010) | $max\left([P_t, t + \Delta t]\right) - min\left([P_t, P_t + \Delta t]\right) > P_{val}$ | A ramp event occurs if the difference between the maximum and minimum power output measured during $\Delta t$ exceeds a pre-set threshold. | Does not characterise rate of change (ramp rate). |
| Zheng and Kusiak (2009)* | $\dfrac{P_t + \Delta t - P_t}{\Delta t} > P_{rr}$ | A ramp event occurs if absolute difference between start and end values of $\Delta t$ and the size of $\Delta t$ itself are greater than pre-set power ramp rate value, $P_{rr}$ | Sensitive to threshold value. |

---

[1] **Model abbreviations:** Time series, TS; Numerical Weather Prediction Model, NWP; Genetic Programming, GP; K-Nearest Neighbour, K-NN; Wavelet Transform, WT; Principal Component Analysis, PCA; Support Vector Machines, SVM; Auto-Regressive, AR; Artificial/Convolutional/Recurrent Neural Network, A/C/RNN; Kernel Density Estimation, KDE; Random Forest, RF; Long Short-Term Memory, LSTM.
**Evaluation Metric abbreviations:** Confusion Matrices, CM; Fuzzy Inference Score, FIS; Mean Absolute (Percentage) Error, MA(P)E; Root Mean Squared Error, RMSE; Ramp Capture, RC; Probability Density Function, PDF; Receiver Operating Characteristic; Forecast Accuracy, FA.

| Bossavy et al. (2010)** | $\left\| P_t^f \right\| > P_{val}$ And: $P_t^f = mean(P_{t+h} - P_{t+h-n}; h = 1,...,n_{nam})$ | Where: $P_t^f$ = filtered version of power signal (transformed using k-order differences in power amplitude) and $n_{nam}$ = number of average power measures to consider (replaces $\Delta t$). Ramp events are identified using the filtered version of the power signal obtained according to the equations. The $n_{nam}$ parameter allows model sensitivity to be tuned to a characteristic ramp event time length that is considered of interest. | Sensitive to threshold value. |
|---|---|---|---|

Table 5. LHB dataset feature description.

| Term | Name | Units | Description |
|---|---|---|---|
| Ws_avg | Wind speed | m/s | 10-min avg from two anemometers on the nacelle of each turbine |
| Wa_avg | Wind direction | degrees (°) | 10-min avg from the wind vane of each turbine |
| P_avg | Active power | kW | 10-min avg from each turbine |
| Ot_avg | Outside temperature | °C | 10-min avg from the wind vane of each turbine |
| **Calculated features** | | | |
| P_tot | Total active power | kW | Sum of P_avg from all turbines |
| %P_rated | Percentage of rated power | % | P_tot divided by 8200 kW*, multiplied by 100 |

The data are recorded at ten-minute intervals between 7 January 2013 (00:20) and 14 December 2017 (04:50). The dataset therefore equates to 1,057,968 observations over a four-year period.

The strategies of the TSO will also depend on how far in advance the ramp is forecast and the availability and the response times of ancillary energy sources. Current network flows and energy demand are also factors that must be considered (Cutler et al., 2007; Martínez-Arellano, 2015) however, these are beyond the scope of this paper. For recent studies modelling the power grid impacts of ramp events, the reader is directed to Veron et al. (2018) and Wang et al. (2016). Using the WT methodology, a ramp function is obtained which provides a non-binary/ continuous index related to ramp intensity at each time step of a given wind power time series (Figure 3).

The wavelets are shifted and dilated versions of a so-called mother wavelet, $\psi^\tau$, and are derived as:

$$\psi^{\tau,\lambda}(t) = \frac{1}{\sqrt{\lambda}} \psi\left(\frac{t-\tau}{\lambda}\right) \qquad (1)$$

where $\tau$ relates to the shift (expressed on a continuous scale between min and max)
$\lambda$ relates to the dilation.

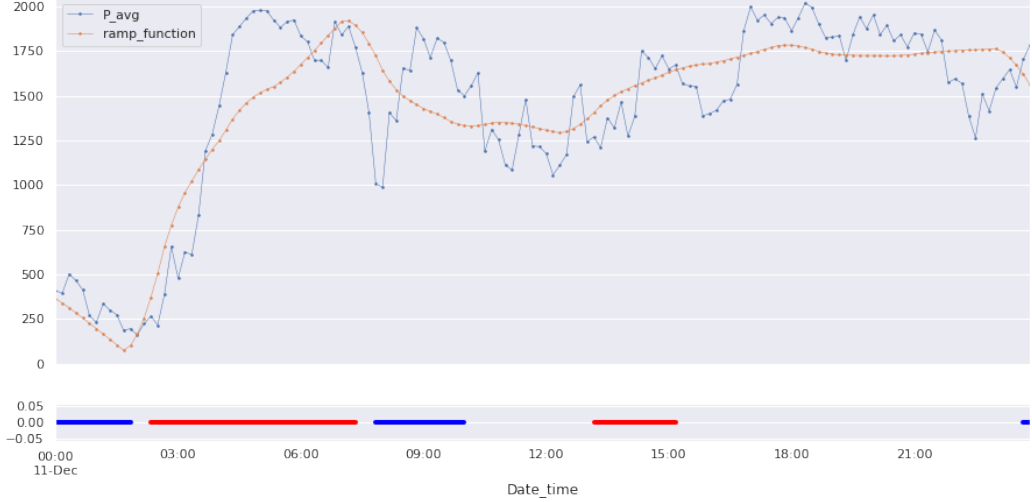

**Figure 3.** Ramp function plot against wind power. The continuous/ non-binary ramp function is plotted in orange. For comparison, the subplot displays a binary ramp function (red = positive ramp, blue = negative ramp, empty space = non-ramp). The difference in granularity between the two ramp definition approaches can be readily observed. For example, depending on the threshold values set, the increase in power output at approx. 17:00 is not captured using a binary ramp definition.

The WT of a time series, $\{y_t\}$, consists of a set of coefficients obtained:

$$W^{\tau,\lambda} = \sum_{-\infty}^{\infty} \psi_t^{\tau,\lambda} \cdot y_t \qquad (2)$$

where $\tau \in Z, \lambda \in Z^+$, and $\psi_t^{\tau,\lambda}$ relates to the wavelet function employed[2].

The fundamental concept of a ramp event is that a certain large gradient is maintained during consecutive time steps of the time series. Therefore, a ramp can be said to show self-similarity. This means that the whole ramp event is similar to a smaller part of it, or in other words, the shape of the event is preserved at different scales. During the period of the ramp event, the wavelet transform, $W^{\tau,\lambda}$, provides increasing coefficients for a broad range of time scales, $\lambda$, since each $W^{\tau,\lambda}$ is related to the gradient at time $t = \tau$, evaluated in a time scale $\Delta t = \lambda$. The increase of $W^{\tau,\lambda}$ with respect to $\lambda$ occurs because the scale is contributed as $\lambda^{-\frac{1}{2}}$ in (1). Hence the gradient is high for both short and long-time scales, where 'long' means close to the length of the overall event. If a period is considered where $\{y_t\}$ exhibits many high-frequency fluctuations, similar coefficients are observed for short scales but not for long scales, in other words, the gradient is no longer self-similar at longer time scales. Finally, if a period is considered where $\{y_t\}$ does not exhibit high gradients, the coefficients $W^{\tau,\lambda}$ would be close to zero for every scale, $\lambda$, considered. This is utilised in the ramp characterisation process by defining the ramp function, $\{R_t\}$, as the sum of the wavelet coefficients $W^{\tau,\lambda}$ at time $t = \tau$ for the interval of scales, $\lambda$, given by $[\lambda_1, \lambda_N]$:

$$R_t(\lambda_1, \lambda_N) = \sum_{\lambda=\lambda_1}^{\lambda_N} W^{\tau,\lambda} \qquad (3)$$

---

[2] The wavelets are defined such that the sign of the coefficient obtained is opposite to that of the ramp event gradient.

In this way, $\{R_t\}$ becomes related to the sharpness of the ramp event because it gathers at each time step the contribution of the gradient evaluated under different timescales. Time scale $\lambda_1$ is the smallest that can be considered, and the minimum possible value is $\lambda_1 = 2$ because at least two values are required to define a gradient. The maximum time window (in terms of time steps) to be evaluated, $\lambda_N$, requires end-user input. According to Gallego et al. (2013), appropriate values of $\lambda$ for hourly wind power time series data are between 5 and 10. Irrespective of this however, the ramp function is not expected to be highly sensitive to small changes in $\lambda$, primarily because the addition of the wavelet transform coefficients across the range of scales from $\lambda_1$ to $\lambda_N$ (2) reduces the impact of any additional scale $\lambda_{N+1}$.

## 4. RESULTS SUMMARY

Untuned model evaluation and ramp performance statistics generated by this study are presented in Table 5. Untuned results are presented to ensure fair comparison between models. Model tuning results will be discussed in a further work. The evaluation criteria deployed are the Mean Absolute Error (MAE) and the Root Mean Square Error (RMSE) for positive, negative and non-ramp events in the TS. The elements to be forecasted are removed from the original time series, and the remaining series consists in the training series. Removed elements are then predicted employing one of the models described in Section 3, which is reported as a model test performance.

Using 10-minutely data, the most accurate model is the LSTM-RNN with a test MAE of 27.34 kW. The LSTM-RNN is followed by ARIMA and ARMA with test MAEs of 49.42 and 50.06 kW respectively. Prophet is the least accurate model with a test MAE of 336.16 kW. Using hourly data, ARMA slightly outperforms ARIMA (test MAEs of 93.1 to 93.26 kW respectively). Otherwise, the ranking remains unchanged.

On average, the LSTM-RNN is able to predict ramp events in the LHB dataset most accurately. ARMA accuracy is generally slightly higher than ARIMA and Prophet is consistently the least accurate, exhibiting a conservative model in its predictions of power output, not venturing very far from data mean. On the other hand, the similarity of the training and test scores for ARIMA indicate that the model is not overfitting the data. ARIMA is again more able than ARMA to predict positive ramp events, but less able to predict non-ramps and negative ramps.

When fitted to 10-minute data, the LSTM-RNN can predict negative ramps almost twice as accurately as it can positive ramps (e.g., 36.28 to 61.06 MAE respectively). The learning curves, however, indicate that the model is overfitting (training performance is much better than test) and is sub-optimally adjusted.

Table 5. Model evaluation and ramp performance statistics generated from this study.

| Model | Data sample rate | Data selection | Lags | Fit time (mm:ss) | Forecast time (mm:ss) | Train RMSE | Test RMSE | Train MAE | Test MAE | Positive ramp acc. (RMSE) | Positive ramp acc. (MAE) | Negative ramp acc. (RMSE) | Negative ramp acc. (MAE) | Non-ramp acc. (RMSE) | Non-ramp acc. (MAE) |
|---|---|---|---|---|---|---|---|---|---|---|---|---|---|---|---|
| ARIMA | 10 min | Univariate | 3 | 01:35 | 00:07 | 79.84 | 87.69 | 46.03 | 49.42 | 123.68 | 83.85 | 129.51 | 86.88 | 79.65 | 43.62 |
| ARMA | 10 min | Univariate | 3 | 04:46 | 00:13 | 79.69 | 87.56 | 46.62 | 50.06 | 124.39 | 84.58 | 127.74 | 85.04 | 79.62 | 44.45 |
| LSTM-RNN | 10 min | Univariate | 1 | 29:34 | 00:03 | - | 45.59 | - | 27.34 | 85.6 | 61.06 | 57.28 | 36.28 | 39.41 | 23.86 |
| LSTM-RNN | 10 min | Multivariate | 1 | 44:02 | 00:03 | - | 44.29 | - | 25.49 | 81.16 | 58.09 | 57.6 | 38.41 | 38.41 | 21.79 |
| Prophet | 10 min | Univariate |  | 08:02 | 07:28 | 381.53 | 454.32 | 281.68 | 336.16 | 757.64 | 595.44 | 540.79 | 388.9 | 411.63 | 310.72 |
| ARIMA | Hourly | Univariate | 3 | 00:42 | 00:03 | 133.16 | 149.1 | 85.11 | 93.26 | 258.57 | 183.22 | 189.31 | 141.91 | 122.75 | 76.36 |
| ARMA | Hourly | Univariate | 3 | 00:41 | 00:04 | 133.2 | 149.2 | 85.23 | 93.1 | 258.99 | 183.47 | 189.19 | 141.47 | 122.81 | 76.17 |
| LSTM-RNN | Hourly | Univariate | 1 | 06:56 | 00:01 | - | 17.78 | - | 16.12 | 22.07 | 20.8 | 21.94 | 20.82 | 16.61 | 15 |
| LSTM-RNN | Hourly | Multivariate | 1 | 07:05 | 00:01 | - | 48.01 | - | 42.95 | 51.73 | 45.91 | 47.72 | 42.74 | 47.6 | 42.65 |
| Prophet | Hourly | Univariate |  | 00:43 | 00:37 | 375.6 | 448.14 | 278.28 | 319.66 | 719.97 | 564.23 | 550.11 | 394.93 | 386.21 | 280.17 |

On the multivariate data, LSTM-RNN fitted to 10-minute is only slightly more accurate than its univariate counterpart (e.g., 25.49 vs 27.34 test MAE respectively). The model can predict negative ramps more accurately than positive ramps (38.41 and 58.09 MAE respectively). The test predictions show that the model often misses extreme high readings.

In general, the three models showed promising prediction accuracy, particularly RNN architectures. Our approach to incorporating wind field variables improved ramp event prediction accuracy.

5. Discussion

This work addresses emerging methods and prospects in the field of ramp event forecasting which is currently an area of active research due its importance in the growing integration of wind energy into electricity markets worldwide (Figure 4).

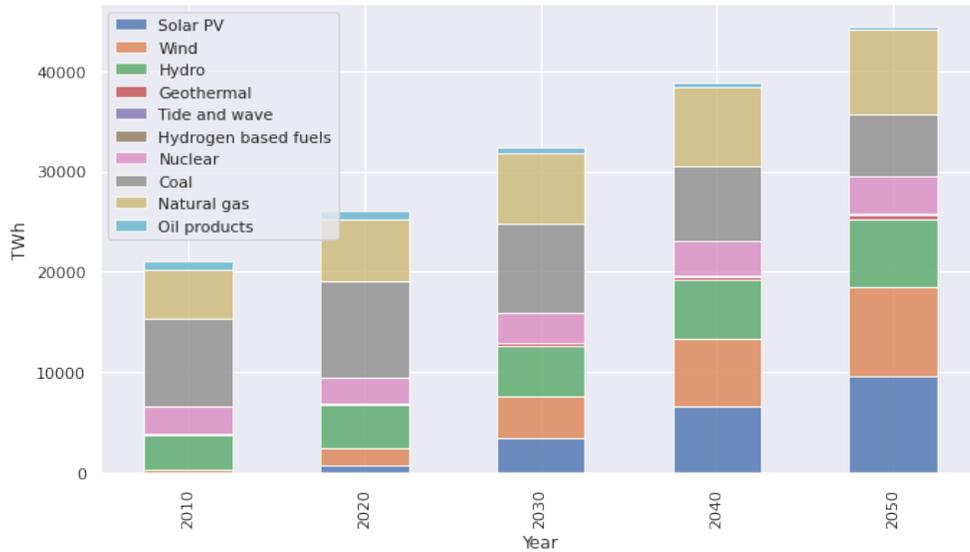

**Figure 4.** Global Electricity Generation Forecast data based on Stated Policies Scenario (data source: International Energy Agency, 2021).

As the wind industry continues to grow, wind turbines are increasingly being installed in large farms where larger turbines provide more cost-effective bulk power to the grid. This will especially be the case in the European offshore sector, where, as part of its Renewable Energy Directive, the European Commission estimates that investments equating to installed capacity of between 240 and 450 GW of offshore wind power will be needed by 2050 to keep global temperature rises below 1.5°C (European Commission, 2021). As wind farm sizes and capacities increase, the effects of sudden changes in wind speed on power output and grid stability become even more significant. This is especially true during ramp events.

While conventional power plants have a reliable production capacity per hour, wind farms are inextricably linked to the natural variability of the wind. Therefore, it is not possible to treat wind power as a conventional energy source. If wind power is to be integrated into the electricity grid as per any other source, it must be considered during the operational procedures of the electricity market, which involves making decisions based on future demand and production. Until such time as new energy storage technologies or alternative innovations are developed, this requires a wind power forecast.

One such alternative innovation currently being developed by turbine manufacturing giant, Siemens Gamesa (SG), is that of green hydrogen: At its Brande Hydrogen pilot test site, SG has coupled an existing onshore wind turbine with an electrolyser stack to investigate the production of hydrogen from the electrolysis of water. The company aims to use commercial volumes of hydrogen to decarbonise refining processes and ammonia production. SG is also using the Brande

Hydrogen site to explore whether integrating new battery technology can contribute to grid stability and help address issues around wind variability. Batteries can store energy that allows electrolysers to run for longer and produce more hydrogen, or they can distribute energy to the grid to help ease supply bottlenecks (Siemens Gamesa Renewable Energy, 2021). Until such alternatives become mainstream, however, management of energy supply through improved forecasting remains a cost-effective solution to the problem of wind power integration.

Using current practices, wind farm costs arising from imbalances between contracted and produced energy are directly proportional to forecast errors (Girard et al., 2013). This relationship was quantified in the case of a Dutch wind farm where improved forecasting was shown to reduce annual (regulatory) costs by 39% (Pinson et al., 2017). Currently, state-of-the-art wind power forecasting systems can achieve a RMSE of 10-15% of total installed capacity over a 36-hour horizon (Martínez-Arellano, 2015; Giebel, G. et al., 2011), thus representing a target for continued development. It is evident then, that improvements in wind power forecasting capability such as those targeted by this work have and will continue to drive down costs of wind power. Cost reductions will play an essential role in improving the competitiveness of wind energy against non-renewable energy sources, enabling a global transition to cleaner fuels, and facilitating climate change mitigation (Figure 4).

Motivated by this, the work presented here provides precursory information that will inform planned further research aiming to improve ramp event forecasting accuracy.

## 6. CONCLUSION AND IMPLICATIONS

We explored the application of advanced machine learning methods into the field of ramp event forecasting which is currently an area of active research due its importance in the growing integration of wind energy into electricity markets worldwide. Wind power is a clean, cheap and abundant source of energy with an unquestionable ability to reduce dependence on fossil fuels. Even among the currently available renewable energy sources, it is believed to be the least harmful to the environment (Abassi et al., 2016).

However, it has also been shown that the variability of wind power and associated errors of forecasting its output can cause short term price rises and price volatilities (Ortega-Vazquez and Kirschen, 2010). Wind farm costs arising from imbalances between contracted and produced energy are directly proportional to forecast errors (Girard et al, 2013).

In this study we examined and evaluated the prediction of wind ramp events using a set of machine learning models, including deep learning, in combination with Wavelet signal transformation approach to LH wind farm dataset. The three examined models, and in particular the RNNs, showed promising accuracy results for wind ramp prediction. This has implications for the planned use of NWP outputs and will require analysis prior to further model development. Possible reasons (and lines of future investigation) may include the resampling and imputation methods used, or the differences in ramp frequencies between the datasets. Despite this, the LSTM model shows the greatest potential for further development. This can be due to its architecture itself; the backwards facing neurons of the LSTM allow it to maintain an internal state of the data it is processing. This fact alone distinguishes it from the other models which are incapable of analysing sequence in the same way. The amplification of this 'memory' with LSTM cells, combined with layering and stacking of neurons into deep networks enables the LSTM-RNN to learn successive layers of increasingly meaningful representations of the data and to better detect long-term dependencies in the sequence being analysed. This makes the LSTM-RNN well suited to the complexities of the LHB dataset, with promising ability to provide feasible forecast of ramp event dynamics.

One of the primary limits of this work is its lack of systematic model tuning. Whilst it could be argued that comparing an ARMA with an ARIMA model is in effect a consideration of model tuning, the Prophet and RNN models potentially offer additional accuracy that has not been fully explored here.

## Acknowledgments

RS and HI would like to thank Alexis Tantet from Laboratoire de Météorologie Dynamique, École Polytechnique (Paris, France) for the initial discussions leading to this study.

Wang, L. et al. (2021) Effective wind power prediction using novel deep learning network: Stacked independently recurrent autoencoder. *Renewable Energy*. 164, pp.642–655.

Wang, Q. et al. (2016) The value of improved wind power forecasting: Grid flexibility quantification, ramp capability analysis, and impacts of electricity market operation timescales. *Applied Energy*. 184, pp.696–713.

Xunta de Galicia (2021) *Xunta de Galicia | Consellería de medio ambiente, territorio e vivenda, MeteoGalicia*2021 [online]. Available from: https://www.meteogalicia.gal/web/inicio.action?request_locale=es [Accessed 12 July 2022].

Yang, B. et al. (2021) State-of-the-art one-stop handbook on wind forecasting technologies: An overview of classifications, methodologies, and analysis. *Journal of Cleaner Production*. 283, p.124628.

Ye, L. et al. (2020) *Combined Approach for Short-Term Wind Power Forecasting Considering Meteorological Fluctuation and Feature Extraction*In: *2020 IEEE/IAS Industrial and Commercial Power System Asia (I CPS Asia)*. 2020 IEEE/IAS Industrial and Commercial Power System Asia (I CPS Asia) 1334–1343.

Yu, C. et al. (2018) A novel framework for wind speed prediction based on recurrent neural networks and support vector machine. *Energy Conversion and Management*. 178, Elsevier, pp.137–145.

Zack, J.W. (2007) *Optimization of wind power production forecast performance during critical periods for grid management*Proceedings of the European Wind Energy Conference EWEC, Milano (IT) 8.

Zhang, Y. et al. (2020a) A new prediction method based on VMD-PRBF-ARMA-E model considering wind speed characteristic. *Energy Conversion and Management*. 203, p.112254.

Zhang, Y. et al. (2020b) Wind Speed Interval Prediction Based on Lorenz Disturbance Distribution. *IEEE Transactions on Sustainable Energy*. [online]. IEEE Transactions on Sustainable Energy 11 (2), pp.807–816.

Zhang, Y. et al. (2018) Wind Speed Prediction of IPSO-BP Neural Network Based on Lorenz Disturbance. *IEEE Access*. IEEE Access 6, pp.53168–53179.

Zhang, Y. et al. (2019) Wind Speed Prediction Research Considering Wind Speed Ramp and Residual Distribution. *IEEE Access*. IEEE Access 7, pp.131873–131887.

Zhao, Y. et al. (2020) Bayesian Network Based Imprecise Probability Estimation Method for Wind Power Ramp Events. *Journal of Modern Power Systems and Clean Energy*. [online]. Journal of Modern Power Systems and Clean Energy pp.1–10.

Zheng, H. and Kusiak, A. (2009) Prediction of Wind Farm Power Ramp Rates: A Data-Mining Approach. *Journal of Solar Energy Engineering*. [online]. 131 (3). Available from: https://doi.org/10.1115/1.3142727 [Accessed 12 July 2022].

Zhou, B. et al. (2021) Short-term prediction of wind power and its ramp events based on semi-supervised generative adversarial network. *International Journal of Electrical Power & Energy Systems*. 125, p.106411.

18